\pdfoutput=1

\documentclass[11pt]{article}

\usepackage[]{ACL2023}

\usepackage{times}
\usepackage{multirow}
\usepackage{latexsym}
\usepackage{graphicx}
\usepackage{colortbl}
\usepackage{amsmath}
\usepackage{amsfonts}
\usepackage{amssymb}
\usepackage{cleveref}
\usepackage{color}
\usepackage{tabularray}
\usepackage{listings}
\usepackage{subcaption}
\usepackage[frozencache,cachedir=.]{minted}
\usepackage{tcolorbox}
\usepackage{lscape}
\usepackage{placeins}

\usepackage{xspace}

\usepackage{color-edits}
\addauthor{gn}{magenta}

\definecolor{Olivine}{rgb}{0.576,0.768,0.49}
\definecolor{BarleyWhite}{rgb}{1,0.949,0.8}
\definecolor{SeaPink}{rgb}{0.917,0.6,0.6}
\definecolor{Cornflower}{rgb}{0.623,0.772,0.909}

\newcommand{\CSM}{OpenAI models\xspace}
\newcommand{\OSM}{LLaMA models\xspace}
\newcommand{\pal}{\textsc{PaL}\xspace}
\newcommand{\COT}[0]{\textsc{CoT}\xspace}

\usepackage[T1]{fontenc}

\usepackage[utf8]{inputenc}

\usepackage{microtype}

\usepackage{inconsolata}
\usepackage{ulem}
\title{Program-Aided Reasoners (\textit{Better}) Know What They Know}

\author{%
  Anubha Kabra\thanks{*Equal contribution}, Sanketh Rangreji$^*$, Yash Mathur$^*$, \\ \textbf{Aman Madaan, Emmy Liu, Graham Neubig} \\
  Language Technologies Institute, Carnegie Mellon University \\
  \texttt{\{anubhak,srangrej,ymathur,amadaan,mengyan3,gneubig\}@andrew.cmu.edu}
}

\begin{document}
\maketitle
\begin{abstract}




Prior work shows that program-aided reasoning, in which large language models (LLMs)  are combined with programs written in programming languages such as Python, can significantly improve accuracy on various reasoning tasks. However, while accuracy is essential, it is also important for such reasoners to ``know what they know'', which can be quantified through the \emph{calibration} of the model. In this paper, we compare the calibration of Program Aided Language Models (\pal{})  and text-based Chain-of-thought (\COT{})  prompting techniques over 5 datasets and 2 model types - \OSM{} and \CSM{}. Our results indicate that \pal{} leads to improved calibration in 75\% of the instances.
Our analysis uncovers that prompting styles that produce lesser diversity in generations also have more calibrated results, and thus we also experiment with inducing lower generation diversity using temperature scaling and find that for certain temperatures, \pal{} is not only more accurate but is also more calibrated than \COT{}.
Overall, we demonstrate that, in the majority of cases, program-aided reasoners \emph{better} know what they know than text-based counterparts.\footnote{Code and data are available at \url{https://github.com/mathuryash5/code-calibrates}.}


\end{abstract}






\section{Introduction}

\begin{figure}[hbt!]
    \centering
    \includegraphics[width= \columnwidth]{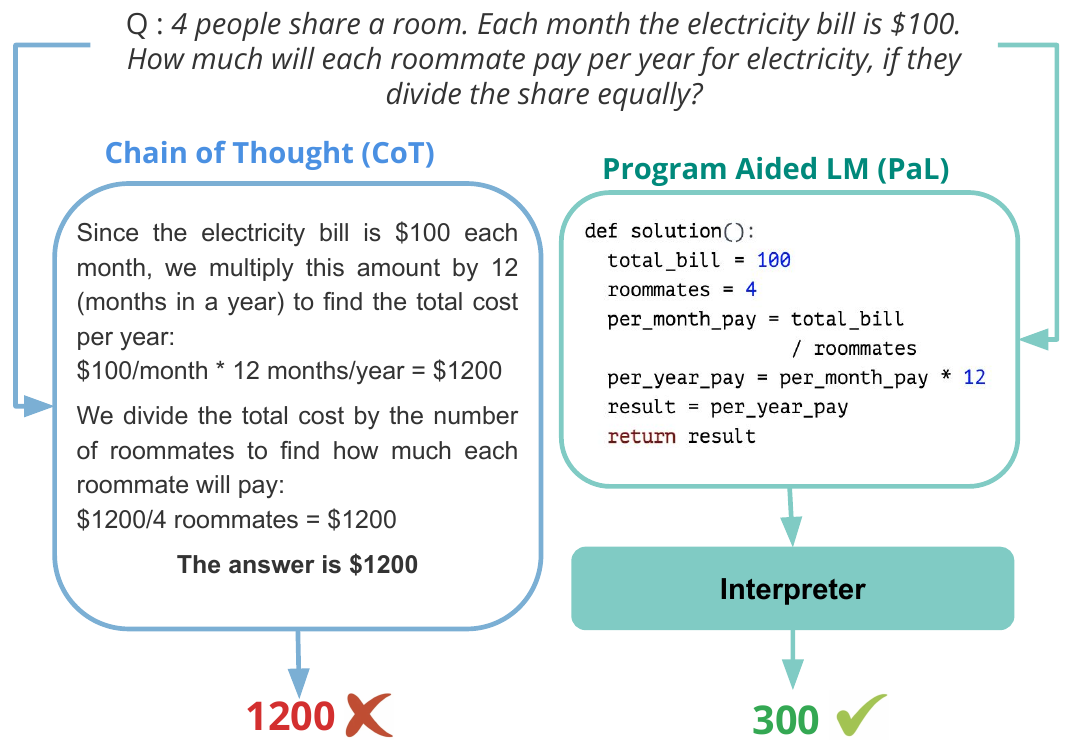}
    \caption{Comparisons of \COT and \pal  outputs. \COT can sometimes generate the correct reasoning chain but fail to derive the correct answer as a final step, \pal fixes this issue by executing generated code to arrive at a deterministic answer.}
    \label{fig:intro_prompting}
\end{figure}

As language models ( LMs )  grow in size and capabilities, several works examine methods to improving their reasoning skills with different styles of prompting \cite{Wei2022ChainOT, Wang2022SelfConsistencyIC, Suzgun2022ChallengingBT, Zhou2022LeasttoMostPE, Yao2023TreeOT}.
One representative method, chain of thought ( \COT )  reasoning \citep{Wei2022ChainOT}, takes inspiration from how humans approach problem-solving -- by breaking down the problem into a sequence of natural language explanations before arriving at a final answer.
Furthermore, prompts that enable problem-solving are not limited to natural language; program-aided language models (\pal); \citet{Gao2022PALPL}  have demonstrated the efficacy of using code (such as Python programs)  as a means of improving the model's reasoning, surpassing the accuracy of conventional chain-of-thought style prompts in some tasks \cite{Madaan2022LanguageMO, lyu2023faithful, Zhang2023ExploringTC, Zhang2023CausalRO}.
An illustration of both methods is shown in \autoref{fig:intro_prompting}.

Currently, most works proposing such methods have been primarily focused on improving accuracy. However, for real-world applications, another highly desirable feature of ML systems is that they should be able to provide \emph{reliable confidence estimates}.
Accurate estimates of model confidence are helpful for many applications, including allowing the model to refrain from providing an answer when uncertain, asking for human intervention in uncertain cases, or providing confidence estimates to a downstream model that consumes the outputs.
The reliability is measured through \emph{calibration}, how a model's confidence in its predictions aligns accurately with real outcomes \citep{guo2017calibration, Jiang2020HowCW, zhao2021calibrate}.
In sum, the previous research has shown, as eloquently stated by \citet{kadavath2022language} ``language models \emph{( mostly ) } know what they know'' --- LLMs are reasonably well calibrated, although some imperfections remain.

In this work, we examine the effect of program-aided reasoning on calibration.
We consider 5 datasets that cover different reasoning tasks and evaluate the performance of both \pal{} and \COT{} style prompting for \CSM{} \cite{openAI} and \OSM{} \cite{touvron2023llama} with respect to accuracy and calibration.
We primarily explore three main research questions :

\begin{itemize}

    \item \textit{ \textbf{RQ 1:} Does program-aided reasoning result in significantly different calibration than text-based \COT{}?}

    \item \textit{ \textbf{RQ 2:} Are the observed trends different across  \CSM{} and \OSM{}? }
    
    \item \textit{\textbf{RQ 3:} Does
    the consistency of LLM generations 
    affect calibration? We examine this by 
    measuring generation diversity and answer space entropy.} 
\end{itemize}

Our results show that program-aided reasoners know what they know \emph{even better} than standard text-based reasoners with \COT.
In particular, on \CSM{}, \pal{} exhibits not only superior accuracy, but also a consistent enhancement in calibration, of about 50\%, over \COT{}. Interestingly, the consistent improvement of calibration is not observed in \OSM{}, but we find that by adjusting the temperature of sampling ( similar to a widely used method of Platt scaling \citep{platt1999probabilistic} ), \pal improves with respect to both accuracy and calibration.
We also conduct a detailed analysis of these observations, and find a correlation between the similarity of the generated chains-of-thoughts or programs and calibration which might help in  explaining these trends.

\section{Preliminaries and Mathematical Formulation}

\subsection{Measuring Calibration}

Calibration refers to the alignment between the predicted probability estimates of a model and their actual correctness or accuracy \cite{Guo2017OnCO}. 
Formally a perfectly calibrated model can be expressed using the following equation, where $X$ is the given input, $Y$ is the true output, the model's output is $\hat{Y}$ and $ P_N (\hat{Y} \mid X) =p $ is the probability, or ``confidence'', over the model's output.

\begin{align}
    P\left (\hat{Y}=Y \mid P_N (\hat{Y} \mid X) =p\right) =p, 
    \forall p \in[0,1]
    \label{eqn:calib}
\end{align}


In essence, Equation~\ref{eqn:calib} conveys that if a perfectly calibrated model makes 100 predictions, and the confidence of each prediction is 0.6 then we expect the accuracy to be also 0.6. 
Nevertheless, the model may exhibit varying confidence levels for each sample. Therefore, it is imperative to calculate calibration across all confidence scores. In practice, we estimate this probability by dividing the predictions into $M$ separate and equally sized interval buckets based on their confidence levels.

We use the expected calibration error (ECE) , a common measure of (lack of)  calibration which is a weighted average of the discrepancy between each bucket’s accuracy and confidence. It is given in \autoref{eq:ece} 



Here $B_m$ is the $\textit{m}$-th bucket that contains samples whose probabilities of predictions fall in the interval $\left (\frac{m-1}{M}, \frac{m}{M} \right] $, where $\frac{\left| B_m \right|}{n}$ is $B_m$'s size relative to all the samples. $\operatorname{acc}\left (B_m\right) $ is the average accuracy of the samples in the $\textit{m}$-th bucket, and $\operatorname{conf}\left (B_m\right) $ is the corresponding average confidence of the samples falling in the $\textit{m}$-th bucket. 

\begin{align}
    \label{eq:ece}
    \sum_{m=1}^M \frac{\left|B_m\right|}{n}\left|\operatorname{acc}\left (B_m\right) -\operatorname{conf}\left (B_m\right) \right| 
\end{align}

Consider a setup where we have buckets with a step size of 0.1. All instances where a model assigns probabilities between 0.4 and 0.5 will be allocated to the bucket $B_4$ or the bucket encompassing probabilities between 0.4 and 0.5. We then calculate the average accuracy for the instances in these buckets along with the average probability/confidence. The absolute difference is multiplied by the proportion of total instances in a bucket. This process is repeated for every bucket and the individual scores are summed up to calculate ECE.

\subsection{Self-consistency as a measure of confidence} 


Self-consistency \cite{Wang2022SelfConsistencyIC} is a technique for natural language reasoning that involves using chain-of-thought prompting to generate multiple paths for reasoning. This process aims to select the most consistent answer by sampling and marginalizing. Here we use a latent variable $Z$ to represent the reasoning chain/programs. $Y$ is the answer that is either extracted in case of \COT{} or obtained after execution in case of \pal{}. We marginalize over $Z$ by taking a majority vote over answers. Thus we rely on majority voting over the answers for obtaining confidence estimates for each sample. 

\begin{figure}[!hbt]
    \centering
    \includegraphics[width= \columnwidth]{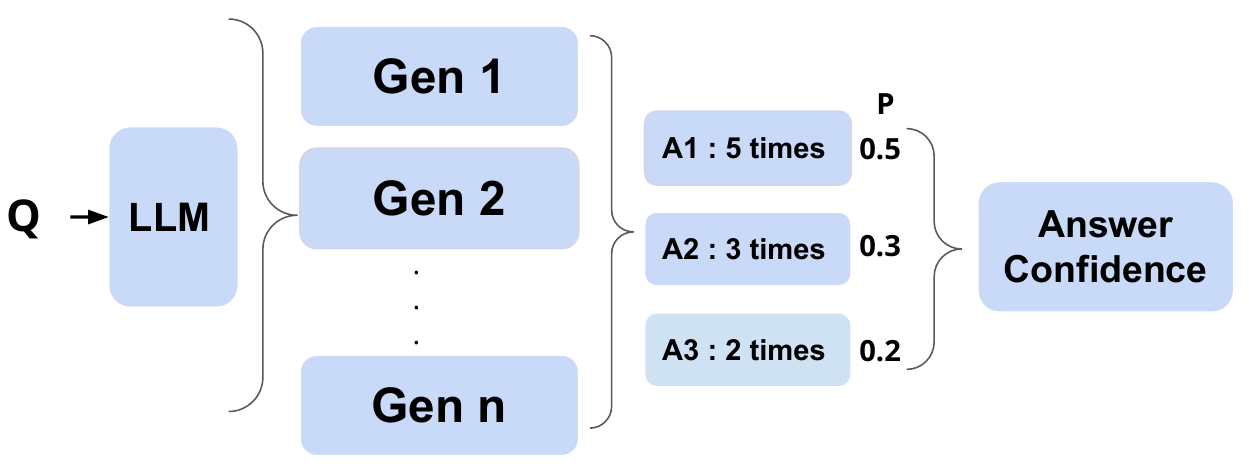}
    \caption{An illustration of obtaining model confidence through majority voting over the answers ($A_1, A_2... A_n$). }
    \label{fig:self-consistency}
\end{figure}

$K$ is a hyperparameter that controls the number of generations  (referenced in equation \ref{eq:self_consistency}). The higher the value of $K$, the better our approximation of the probability of each sample. An overview of this process is shown in Figure \ref{fig:self-consistency}.

\begin{equation}
\label{eq:self_consistency}
P (\hat{Y}_0|Z_0)=\frac{1}{K} \sum_{i=0}^K \mathbb{I}\left\{\hat{Y}_i=\hat{Y}_0\right\}
\end{equation}

\citet{Wang2022SelfConsistencyIC}  and \citet{Xiong2023CanLE}
suggest that self-consistency can be an effective way to elicit confidence from models. Hence, given the lack of per-token log probabilities in closed LMs like \texttt{gpt-3.5-turbo} and \texttt{text-davinci-003}, we adopt self-consistency as a proxy measure for calibration.

\subsection{Similarity and Answer Entropy}


In addition to empirically evaluating the impact on accuracy and calibration, we conduct a qualitative analysis of the reasoning chains (which can be thought of as the latent variable $Z$ described previously). Here, we observe a consistent pattern: the correct answers corresponding to a question were often associated with similar generations. This observation led us to hypothesize that this phenomenon could be attributed to the fact that there are numerous ways in which solutions can be incorrect, whereas correct solutions tend to exhibit more uniform behaviour \cite{li2022competition}. To empirically assess this hypothesis, we employed sentence embeddings generated from the \textit{all-MiniLM-v6} model to compute the average similarity among the generations which is equivalent to calculating similarity over latent variables $Z$. 

Furthermore, to gain deeper insights into the relationship between similarity in generations and corresponding answers, we also compute the entropy $H (A)$ of the answer space where $P (a_i)$ refers to the probability of the $i^{th}$ answer in K answers obtained by extraction or program execution for a given sample.

\begin{equation}
H (A) = -\sum_{i=1}^{K} P (a_i) \cdot \log_2 P (a_i)
\end{equation}

This allowed us to investigate whether the observed similarity in the latent variable space $Z$  leads to a lower entropy within the answer space. These quantitative measures were useful in gaining insights into why specific dimensions yielded more favourable evaluation metrics.

\begin{table*}
\centering
\resizebox{\linewidth}{!}{%
\begin{tblr}{
  hline{1-2,7} = {-}{},
}
Dataset            & Category        & \# Samples & Example                                                                                                                                    \\
GSM8K \cite{Cobbe2021TrainingVT}             & Arithmetic  & 1319       & {Q: A robe takes 2 bolts of blue fiber and half that much white fiber.\\How many bolts in total does it take?\\A: 3}                       \\
GSM8K Hard \cite{Gao2022PALPL}        & Arithmetic  & 1319       & {Q: A robe takes 2287720 bolts of blue fiber and half that much white fiber.~\\~How many bolts in total does it take?\\A: 3431580}         \\
Date Understanding  \cite{suzgun2022challenging}& Symbolic    & 360        & {Q: Yesterday was April 30, 2021. \\What is the date today in MM/DD/YYYY?\\A: 05/01/2021}                                                  \\
Object Counting \cite{suzgun2022challenging}    & Algorithmic & 250       & {Q: I have three couches, a lamp, a stove, a table, a fridge, \\and a microwave. How many objects do I have?\\A: 8}                        \\
Repeat Copy \cite{suzgun2022challenging}       & Algorithmic & 32         & {Q: say python twice and data once, and then repeat all of this three times.\\A: python python data python python data python python data} 
\end{tblr}
}
\caption{ Datasets with their examples and categories.}
\label{tab: datasets}
\end{table*}

\section{Experimental Design}

\subsection{Models}
We compare the calibration and accuracy of two different prompting strategies - CoT and PaL on an equal number of closed-source and open-source models. The open source models used in experimentation are \texttt{LLaMA2-13B, LLaMA2-70B}
and the closed-source models are \texttt{gpt-3.5-turbo, text-davinci-003} \cite{brown2020language}. 
It should be noted that all models have received some form of supervision from code during pre-training \cite{openAI, touvron2023llama}, in addition to being primarily trained on text.

\subsection{Hyperparameters}

For our experiments, we set temperature (T)  as 1.0 and the probability (p)  for nucleus sampling \cite{holtzman2020curious} as 1.0. 
Selecting a temperature of 1.0 enables direct sampling from the model as there is no scaling of probabilities involved, as seen from Equation \ref{eq:temperature}. Here, $z_i$ refers to the logit for the $i$th token generated and $N$ is the size of the vocabulary.  

\begin{equation}
\label{eq:temperature}
\sigma\left (z_i\right)=\frac{e^{\frac{z_i}{T}}}{\sum_{j=0}^N e^{\frac{z_j}{T}}}
\end{equation}

For each sample in a dataset, we set the number of generations ($K$)  as 10. For each generation, we set the maximum number of tokens (input + output) at 1024.

\subsection{Tasks}

We examined reasoning tasks encompassing several challenges that include arithmetic, algorithmic, and symbolic reasoning. We use five datasets that cover these different kinds of reasoning tasks. The arithmetic reasoning datasets include \textit{GSM8K} \cite{Cobbe2021TrainingVT} and \textit{GSM8K Hard} \cite{Gao2022PALPL}. The algorithmic reasoning tasks include \textit{Object-Counting} \cite{suzgun2022challenging} and \textit{Repeat-Copy }\cite{suzgun2022challenging}. We used \textit{Date-Understanding} as a Symbolic Reasoning Dataset \cite{suzgun2022challenging}. Specific information about the datasets used can be found in \autoref{tab: datasets}.


\subsection{Prompt Design}
We provide all models with natural language chain-of-thought  (CoT)  prompts and code-based Program-Aided Language Model (PaL)  prompts. For datasets where CoT prompts are available in their original form, we use them as presented in the original paper \cite{Wei2022ChainOT}. For other datasets, we modify these prompts to suit the specific task while maintaining their original format. For PaL prompts we use and adapt the code-prompts provided in \cite{Gao2022PALPL}. 
The prompts can be seen in Appendix Section \ref{sec:prompts}.

\section{Results}
\begin{table*}[hbt]
\centering

\resizebox{\textwidth}{!}{%
\begin{tabular}{c|cc|cc|cc|cc|cc|cc}
\hline
Name                              & Score & Model & \multicolumn{2}{c}{GSM8K} & \multicolumn{2}{c}{Object-Counting} & \multicolumn{2}{c}{Repeat-Copy} & \multicolumn{2}{c}{Date-Understanding} & \multicolumn{2}{c}{GSM8K Hard} \\ \hline
                                  &       &       & CoT         & PaL          & CoT              & PaL              & CoT            & PaL            & CoT                & PaL               & CoT            & PaL
                                  \\
                                  \hline
\multirow{3}{*}{\texttt{LLaMA2-70B} }      & ECE ($\downarrow$)     & LLaMA   & \cellcolor{blue!10}0.19       & \cellcolor{blue!20}0.07        & \cellcolor{blue!10}0.17            &\cellcolor{blue!20} 0.14            & \cellcolor{blue!20}0.18          &\cellcolor{blue!10} 0.23          & \cellcolor{blue!20}0.09              & \cellcolor{blue!10}0.18            & \cellcolor{blue!10}0.07          &\cellcolor{blue!20} 0.03          \\
                                  & ACC ($\uparrow$)     & LLaMA   & \cellcolor{blue!10}59.28      & \cellcolor{blue!20}63.91       & \cellcolor{blue!10}76.00               & \cellcolor{blue!20}92.40             & \cellcolor{blue!10}40.62          & \cellcolor{blue!20}71.88          & \cellcolor{blue!10}66.66             & \cellcolor{blue!20}70.18           & \cellcolor{blue!10}21.45         & \cellcolor{blue!20}40.62          \\
                                  & SIM ($\uparrow$)      & LLaMA   & \cellcolor{blue!10}72.20       & \cellcolor{blue!20}92.40        & \cellcolor{blue!10}94.43            & \cellcolor{blue!20}94.72             & \cellcolor{blue!10}87.10           & \cellcolor{blue!20}90.58           & \cellcolor{blue!20}86.87               & \cellcolor{blue!10}82.15               & \cellcolor{blue!10}92.28          & \cellcolor{blue!20} 74.32          \\ 
                                  & ENT ($\downarrow$)      & LLaMA   &\cellcolor{blue!10} 2.24       & \cellcolor{blue!20}1.92        & \cellcolor{blue!10}1.00            &\cellcolor{blue!20} 0.76            & \cellcolor{blue!20}1.93           & \cellcolor{blue!10}2.00           & \cellcolor{blue!20}1.44               & \cellcolor{blue!10}1.54              & \cellcolor{blue!10}2.85           &\cellcolor{blue!20} 2.17           \\ \cline{2-13} 
\multirow{3}{*}{\texttt{LLaMA2-13B}}        & ECE ($\downarrow$)     & LLaMA   & \cellcolor{blue!20}0.06       & \cellcolor{blue!10}0.08        & \cellcolor{blue!10}0.08            & \cellcolor{blue!20}0.06            & \cellcolor{blue!20}0.11          & \cellcolor{blue!10}0.17          & \cellcolor{blue!10}0.06              & \cellcolor{blue!20}\textbf{0.05  }           & \cellcolor{blue!20}0.12          &\cellcolor{blue!10} 0.14          \\
                                  & ACC ($\uparrow$)     & LLaMA   & \cellcolor{blue!10}27.0        & \cellcolor{blue!20}34.34        & \cellcolor{blue!10} 56.4             & \cellcolor{blue!20}81.6             &  \cellcolor{blue!10} \cellcolor{blue!10}34.37          & \cellcolor{blue!20}53.12          &  \cellcolor{blue!10}48.24              & \cellcolor{blue!20}50.41             &  \cellcolor{blue!10}6.67           & \cellcolor{blue!20}25.55          \\
                                  & SIM ($\uparrow$)      & LLaMA   & \cellcolor{blue!10}76.6       & \cellcolor{blue!20}93.3        & \cellcolor{blue!10}93.2            & \cellcolor{blue!20}95.3            & \cellcolor{blue!20}89.8          & \cellcolor{blue!10}88.6          & \cellcolor{blue!10}79.5              & \cellcolor{blue!20}84.2             & \cellcolor{blue!10}74.0           &\cellcolor{blue!20} 92.32          \\
                                  & ENT ($\downarrow$)      & LLaMA   & \cellcolor{blue!10}2.83       & \cellcolor{blue!20}2.49        & \cellcolor{blue!10}1.52            & \cellcolor{blue!20}0.85            & \cellcolor{blue!20}2.43          & \cellcolor{blue!10}2.47          & \cellcolor{blue!10}2.23              & \cellcolor{blue!20}2.06             &\cellcolor{blue!20} 2.42           & \cellcolor{blue!10}3.06          \\ \cline{2-13} 

\multirow{3}{*}{\texttt{text-davinci-003}} & ECE ($\downarrow$)     & OpenAI   & \cellcolor{blue!10}0.04       &\cellcolor{blue!20} 0.03       & \cellcolor{blue!10}0.29           &\cellcolor{blue!20} \textbf{0.02}          & \cellcolor{blue!10}0.20          & \cellcolor{blue!20}\textbf{0.06 }        & \cellcolor{blue!10}0.19            & \cellcolor{blue!20}0.11           & \cellcolor{blue!10}0.15         & \cellcolor{blue!20}0.07         \\
                                  & ACC ($\uparrow$)     & OpenAI   & \cellcolor{blue!10}65.65          & \cellcolor{blue!20}76.49        & \cellcolor{blue!10}59.21            & \cellcolor{blue!20}\textbf{98.00 }              & \cellcolor{blue!10}67.23          & \cellcolor{blue!20}\textbf{93.75 }         & \cellcolor{blue!10}60.70               & \cellcolor{blue!20}72.35             & \cellcolor{blue!10}23.95          & \cellcolor{blue!20}\textbf{71.27}          \\
                                  & SIM ($\uparrow$)      & OpenAI   & \cellcolor{blue!10}90.5            & \cellcolor{blue!20}97.8         & \cellcolor{blue!10}99.1            & \cellcolor{blue!20}99.8            &  \cellcolor{blue!10}96.2              & \cellcolor{blue!20}98.2          & \cellcolor{blue!10}92.4              & \cellcolor{blue!20}97.4             & \cellcolor{blue!10}89.8          & \cellcolor{blue!20}97.9          \\
                                  & ENT ($\downarrow$)      & OpenAI   & \cellcolor{blue!10}1.27           & \cellcolor{blue!20}0.79         & \cellcolor{blue!10}0.36            &\cellcolor{blue!20} 0.02            &  \cellcolor{blue!10}1.38              & \cellcolor{blue!20}0.44          & \cellcolor{blue!10}0.71              &\cellcolor{blue!20} 0.64             & \cellcolor{blue!10}2.31         & \cellcolor{blue!20}0.81          \\ \cline{2-13}
\multirow{3}{*}{\texttt{gpt-3.5-turbo}}    & ECE ($\downarrow$)     & OpenAI   &\cellcolor{blue!10} 0.05       &\cellcolor{blue!20} \textbf{0.03}      & \cellcolor{blue!10}0.38           & \cellcolor{blue!20}0.03           &\cellcolor{blue!10} 0.18         & \cellcolor{blue!20}0.16          & \cellcolor{blue!10}0.17              &\cellcolor{blue!20} 0.13            & \cellcolor{blue!10}0.13         & \cellcolor{blue!20}\textbf{0.05}         \\
                                  & ACC ($\uparrow$)     & OpenAI   & \cellcolor{blue!20}\textbf{84.00 }         & \cellcolor{blue!10}82.40         & \cellcolor{blue!10}82.40             & \cellcolor{blue!20}97.20             & \cellcolor{blue!10}56.25          & \cellcolor{blue!20}68.75          & \cellcolor{blue!10}61.51              &\cellcolor{blue!20} \textbf{77.23 }            & \cellcolor{blue!10}55.21          & \cellcolor{blue!20}62.91          \\
                                  & SIM ($\uparrow$)      & OpenAI   & \cellcolor{blue!10}94.40       & \cellcolor{blue!20}97.80        & \cellcolor{blue!20}99.10            & \cellcolor{blue!10}98.60            & \cellcolor{blue!10}97.70          & \cellcolor{blue!20}97.90           & \cellcolor{blue!10}95.3              & \cellcolor{blue!20}97.6             & \cellcolor{blue!10}90.60          &\cellcolor{blue!20} 95.40          \\ 
                                  & ENT ($\downarrow$)      & OpenAI   & \cellcolor{blue!10}0.57       & \cellcolor{blue!20}0.49        & \cellcolor{blue!10}0.59            &\cellcolor{blue!20} 0.048            & \cellcolor{blue!10}1.15          &\cellcolor{blue!20} 0.35          & \cellcolor{blue!10}0.50              &\cellcolor{blue!20} 0.36             & \cellcolor{blue!20}1.65          &\cellcolor{blue!10} 2.43         \\ \hline
\end{tabular}
}
\caption{Comparison of Expected Calibration Error ({ECE ($\downarrow$)  }) , Accuracy ({ACC ($\uparrow$)  }) , Cosine Similarity ({SIM ($\uparrow$)  })  and Answer Entropy ({ENT ($\downarrow$)  })  across datasets. The darker  \colorbox{blue!20}{blue} shade highlights better performing prompting technique.}
\label{table:result}
\end{table*}

\begin{figure*}[h]
    \centering
    \includegraphics[width= \textwidth]{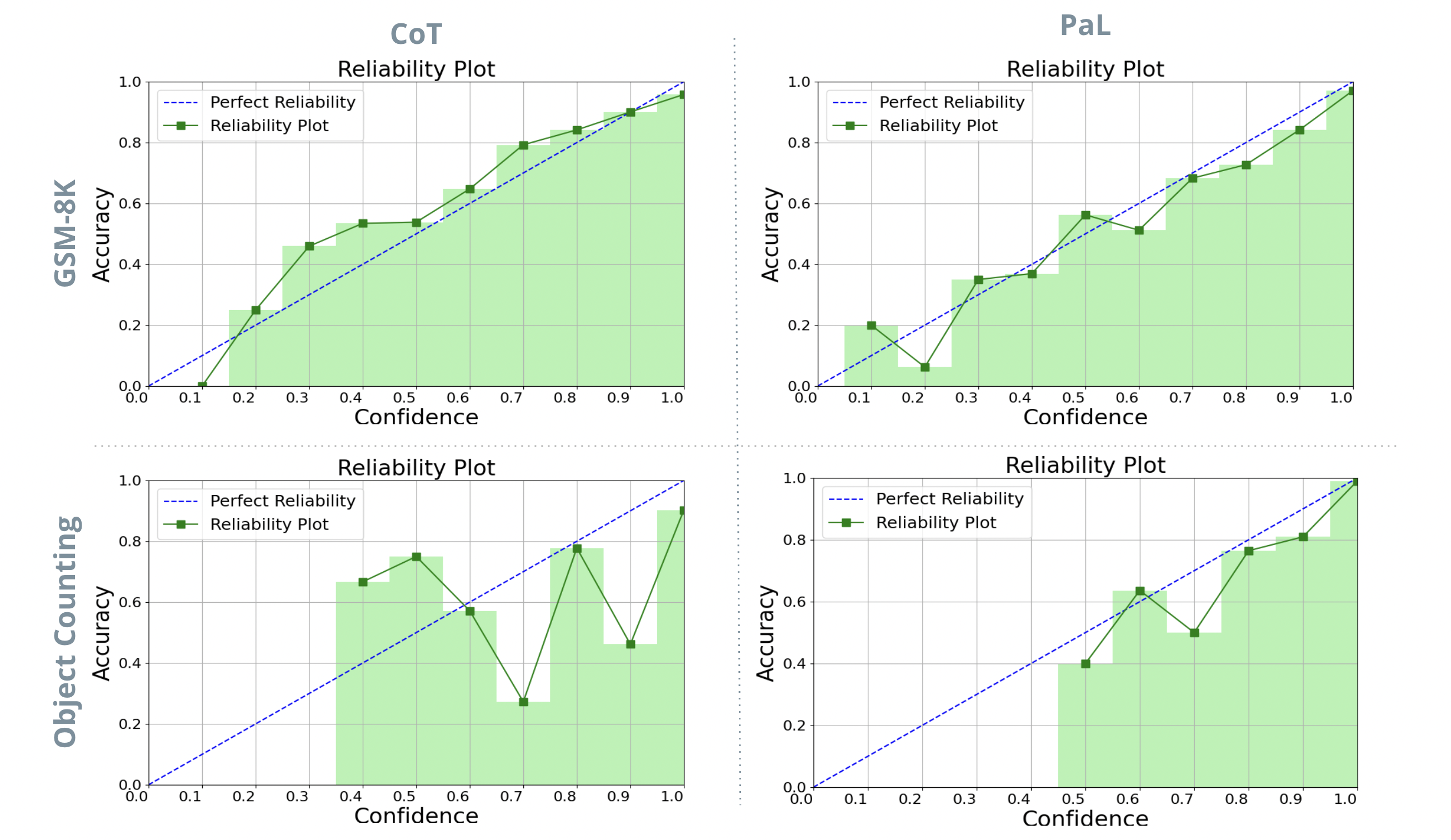}
    \caption{Reliability Plots for various kinds of structured reasoning tasks for the model \texttt{gpt-3.5-turbo }. The x-axis represents confidence and the y-axis represents accuracy. }
    \label{fig:reliability}
\end{figure*}

We investigate two model types: \CSM{} and \OSM{} along with the two different prompting strategies - \pal{} and \COT{}. 
\subsection{Effect of prompting style on Calibration}
In this section, we look at the first two \textit{RQs}: \\
\textit{ \textbf{RQ 1:} Does one prompting style result in significantly better calibration than the other?} \\\textit{ \textbf{RQ 2:} Are the observed calibration trends different across  \CSM{} and \OSM{}? }

Table \ref{table:result} shows results for \CSM{}, in which we can see that \pal{} prompting improves both calibration and accuracy across all datasets. We see approximately $50\%$ relative reduction in calibration error and an average improvement of 18.42\% in accuracy.  In Figure \ref{fig:reliability} we show reliability diagrams, an illustration of the bucket values from Equation \ref{eq:ece}. These provide an illustration of improved calibration, with the reliability curves for \pal{} prompting consistently aligning closer to the ideal reliability curve as compared to \COT{} across datasets. While \pal{} shows a notable gain of 14.83\% in accuracy across all datasets for \OSM{}, it shows better calibration in only half of our settings. Overall for both \CSM{} and \OSM{}, we observe that \pal{} leads to better calibration than \COT{} for 75\% of the settings. 

\begin{table}[hbt!]
\resizebox{\columnwidth}{!}{%
\begin{tabular}{c|ccc}
\hline
Model Type                                                                       & \OSM{}         & \CSM{}         & Both        \\ \hline
\begin{tabular}[c]{@{}c@{}}Fixed Effect\\ (ECE vs Prompting Style) \end{tabular} & PAL : -0.010 & PAL : -0.103 & PAL : -0.067 \\ \hline
p-value                                                                     & 0.961       & 0.000       & 0.002       \\ \hline
\end{tabular}
}
\caption{Statistical analysis using \textit{mixed-LM}, keeping ECE vs Prompting Style as a fixed effect and accuracy as a random effect. }
\label{mixed-lm-results}
\end{table}

\paragraph{Effect of \pal{} on calibration controlling for accuracy}

One reasonable hypothesis is that \pal{} is mainly improving calibration because it achieves higher accuracy, and more accurate models can be better calibrated.
`To examine this hypothesis, we conduct statistical analysis using \emph{mixed linear models} \citep{mclean1991unified}, which allow us to consider the significance of varying the prompting strategy while controlling for accuracy as a confounding factor.


Upon analyzing the results in Table \ref{mixed-lm-results}, we observe that, when treating the prompting style as a fixed effect,  \pal{} exhibits a negative coefficient of -0.103 (p=0.0)  for \CSM{} which is statistically significant with a threshold of p=0.05. This implies that \pal{} contributes to the reduction in ECE, and has a positive impact on calibration.
On the contrary, for \OSM{}, we did not find that \pal{} had a statistically significant effect on ECE after controlling for accuracy.
Across \OSM{} and \CSM{}, \pal{} has a statistically significant (p=0.02)  correlation of -0.067 with ECE, indicating that \pal{} helps increase calibration on the whole even when controlling for accuracy.

To summarize, we see that \pal{} prompting has better calibration than \COT{} prompting (\textit{\textbf{--RQ1}}) . While \pal{} has improved calibration in all settings for \CSM{}, this trend is less consistent for \OSM{} (\textit{\textbf{--RQ2}}) .

\subsection{Effect of generation diversity on calibration}

In this section, we look at the third research question: \textit{\textbf{RQ 3:} Does the consistency of LLM generations affect calibration?} 


Qualitative analysis of the generations reveals that \pal{} generations adhere to a consistent structure that divides the problem-solving process into three distinct parts. This is depicted in Figure \ref{fig:codestructure}. In the first part, the model initializes the variables and sets up their initial values required for the calculation. This part is straightforward due to syntactic constraints and therefore remains largely similar across generations. In the second part, the model generates the required logic by manipulating variables, applying formulas, and utilizing various operations to derive the desired result. Finally, in the third part, the model generates the answer by assigning the calculated value or result to a variable and returning it, which again doesn't vary much across generations. Hence, the diversity of the generation is mostly limited to the second part making code more constrained in its generation space compared to text. Hence we observe a \textbf{standardized structure in the code} generated by language models with PaL prompts. 

\begin{figure}[t!]
\begin{minted}[linenos=false, fontsize=\small]{python}
def solution ()  :
    # Part 1: Initialize
    num_glasses = 16
    first_glass_price = 5
    second_glass_discount = 0.6

    # Part 2: Calculate
    second_glass_price = first_glass_price *
                         second_glass_discount
    pair_price = first_glass_price +
                 second_glass_price
    num_pairs = num_glasses // 2
    total_cost = num_pairs * pair_price

    # Part 3: Result Generation
    result = total_cost
    return result
\end{minted}
\caption{Typical output structure with PaL}
\label{fig:codestructure}
\end{figure}


\paragraph{Lower generation diversity and answer entropy observed in prompting strategy with better calibration} To quantitatively analyze if code-based generations have lower generation diversity and hence lead to a narrower answer space, we computed aggregated cosine similarity scores for all the generations and entropy over the answer space. 
 
 For \CSM{}, we note that the cosine similarity scores with \pal{} are higher than the corresponding scores for \COT{}. This observation suggests that, from a semantic perspective code-based generations display a higher degree of similarity. Moreover, the answer entropy for \pal{} is lower than \COT{}. This implies that similar generations that cluster together in the semantic space \cite{li2022competition}, also converge to the similar solution space. This leads to lower uncertainty in the probability distribution of the answer space and hence lower entropy. From Table \ref{table:result}, we thus can see that \pal{} helps produce similar generations that converge to the same answer space, which is also consistently correct. Hence, achieving better performance and providing more confidence in its predictions. 

For \OSM{}, we don't see this trend of \pal{} having higher generation similarity and lower answer entropy for all datasets. However, for almost all settings for \OSM{} and \CSM{}, the prompting strategy that produces more similar generations and lower answer entropy is also more calibrated.

To summarize, it is evident that lower generation diversity and lower answer entropy are correlated with higher calibration. (\textit{\textbf{--RQ3}}) 

\begin{table*}[hbt!]
\centering
\resizebox{0.95\textwidth}{!}{%
\begin{tabular}{cc|cc|cc|cc|cc|cc}
\hline
Temp                 &     & \multicolumn{2}{c}{GSM8K} & \multicolumn{2}{c}{Object-Counting} & \multicolumn{2}{c}{Repeat-Copy} & \multicolumn{2}{c}{Date-Understanding} & \multicolumn{2}{c}{GSM8K Hard} \\ \hline
                     &     & CoT          & PaL         & CoT              & PaL              & CoT            & PaL            & CoT                & PaL               & CoT            & PaL            \\ \hline
\multirow{4}{*}{0.7} & ECE & \cellcolor{blue!10}0.101        & \cellcolor{blue!20}0.07       & \cellcolor{blue!10}0.06            & \cellcolor{blue!20}0.03            & \cellcolor{blue!10}0.14          & \cellcolor{blue!20}0.12          & \cellcolor{blue!10}0.12              & \cellcolor{blue!20}0.09             & \cellcolor{blue!10}0.18          & \cellcolor{blue!20}0.03           \\
                     & ACC & \cellcolor{blue!10} 66.03           & \cellcolor{blue!20} 67.9        & \cellcolor{blue!10} 77.6             & \cellcolor{blue!20} 93.2             & \cellcolor{blue!10} 53.1           & \cellcolor{blue!20} 75.0             & \cellcolor{blue!10} 74.5               & \cellcolor{blue!20} 76.42              & \cellcolor{blue!10} 27.14           & \cellcolor{blue!20} 52.91           \\
                     & SIM & \cellcolor{blue!10} 85.07        & \cellcolor{blue!20} 97.47       & \cellcolor{blue!10} 98.53            & \cellcolor{blue!20} 99.42            & \cellcolor{blue!10} 93.78          & \cellcolor{blue!20} 94.81          & \cellcolor{blue!10} 89.62              & \cellcolor{blue!20} 96.16             & \cellcolor{blue!10} 83.28          & \cellcolor{blue!20} 97.29         \\
                     & ENT & \cellcolor{blue!10}1.60          & \cellcolor{blue!20}1.48       & \cellcolor{blue!10}0.55            & \cellcolor{blue!20}0.21           & \cellcolor{blue!10}1.46           & \cellcolor{blue!20}1.35          & \cellcolor{blue!10}0.88              & \cellcolor{blue!20}0.80             & \cellcolor{blue!10}2.43          & \cellcolor{blue!20}1.72          \\ \hline
\multirow{4}{*}{0.5} & ECE & \cellcolor{blue!10}0.049        & \cellcolor{blue!20}0.036       & \cellcolor{blue!10}0.103            & \cellcolor{blue!20}0.059           & \cellcolor{blue!10}0.112          & \cellcolor{blue!20}0.075          & \cellcolor{blue!10}0.114             & \cellcolor{blue!20}0.063             & \cellcolor{blue!10}0.139          & \cellcolor{blue!20}0.104          \\
                     & ACC & \cellcolor{blue!10} 66.94        & \cellcolor{blue!20} 67.24       & \cellcolor{blue!10}77.23             & \cellcolor{blue!20}92.4             & \cellcolor{blue!10}59.3           & \cellcolor{blue!20}68.75           & \cellcolor{blue!10}73.44               & \cellcolor{blue!20}77.2             & \cellcolor{blue!10}27.7           & \cellcolor{blue!20}51.63          \\
                     & SIM & \cellcolor{blue!10}88.69        & \cellcolor{blue!20}98.25       & \cellcolor{blue!10}99.17            & \cellcolor{blue!20}99.85            & \cellcolor{blue!20}97.09          & \cellcolor{blue!10}96.81          & \cellcolor{blue!10}92.49              & \cellcolor{blue!20}97.97             & \cellcolor{blue!10}87.65          & \cellcolor{blue!20}98.2         \\
                     & ENT & \cellcolor{blue!10}1.35         & \cellcolor{blue!20}1.19       & \cellcolor{blue!10}0.39            & \cellcolor{blue!20}0.12             & \cellcolor{blue!10}1.09          & \cellcolor{blue!20}0.99          & \cellcolor{blue!10}0.60              & \cellcolor{blue!20}0.52            & \cellcolor{blue!10}2.18         & \cellcolor{blue!20}1.39         \\ \hline
\multirow{4}{*}{0.3} & ECE & \cellcolor{blue!20}0.057        & \cellcolor{blue!10}0.097       & \cellcolor{blue!10}0.140             & \cellcolor{blue!20}0.064            & \cellcolor{blue!10}0.194          & \cellcolor{blue!20}0.113          & \cellcolor{blue!10}0.153              & \cellcolor{blue!20}0.139             & \cellcolor{blue!10}0.230           & \cellcolor{blue!20}0.206          \\
                     & ACC & \cellcolor{blue!20}64.89         & \cellcolor{blue!10}63.38        & \cellcolor{blue!10}78.8             & \cellcolor{blue!20}91.2             & \cellcolor{blue!10}53.12           & \cellcolor{blue!20}71.87           & \cellcolor{blue!10}72.62               & \cellcolor{blue!20}76.42              & \cellcolor{blue!10}26.16           & \cellcolor{blue!20}49.28           \\
                     & SIM & \cellcolor{blue!10}91.91        & \cellcolor{blue!20}98.75       & \cellcolor{blue!10}99.51            & \cellcolor{blue!20}99.94            & \cellcolor{blue!10}97.73          & \cellcolor{blue!20}98.27          & \cellcolor{blue!10}95.18              & \cellcolor{blue!20}99.02              & \cellcolor{blue!10}91.14          & \cellcolor{blue!20}98.75          \\
                     & ENT & \cellcolor{blue!10}1.087        & \cellcolor{blue!20}0.960        & \cellcolor{blue!10}0.238            & \cellcolor{blue!20}0.056            & \cellcolor{blue!10}0.780           & \cellcolor{blue!20}0.504          & \cellcolor{blue!10}0.420               & \cellcolor{blue!20}0.317             & \cellcolor{blue!10}1.866          & \cellcolor{blue!20}1.076          \\ \hline
\multirow{4}{*}{0.1} & ECE & \cellcolor{blue!20}0.219        & \cellcolor{blue!10}0.257       & \cellcolor{blue!10}0.188           & \cellcolor{blue!20}0.07             & \cellcolor{blue!10}0.278          & \cellcolor{blue!20}0.156          & \cellcolor{blue!10}0.233              & \cellcolor{blue!20}0.176             & \cellcolor{blue!10}0.418          & \cellcolor{blue!20}0.380           \\
                     & ACC & \cellcolor{blue!20}58.6         & \cellcolor{blue!10}58.37       & \cellcolor{blue!10}77.2             & \cellcolor{blue!20}90.4             & \cellcolor{blue!10}53.12           & \cellcolor{blue!20}68.75           & \cellcolor{blue!10}69.91               & \cellcolor{blue!20}78.32              & \cellcolor{blue!10}23.5           & \cellcolor{blue!20}45.87           \\
                     & SIM & \cellcolor{blue!20}95.79        & \cellcolor{blue!10}99.37       & \cellcolor{blue!10}99.82            & \cellcolor{blue!20}99.98            & \cellcolor{blue!10}99.28          & \cellcolor{blue!20}99.64          & \cellcolor{blue!10}98.21              & \cellcolor{blue!20}99.68             & \cellcolor{blue!10}95.31          & \cellcolor{blue!20}99.35          \\
                     & ENT & \cellcolor{blue!10}0.661        & \cellcolor{blue!20}0.526       & \cellcolor{blue!10}0.085            & \cellcolor{blue!20}0.026            & \cellcolor{blue!10}0.288          & \cellcolor{blue!20}0.173          & \cellcolor{blue!10}0.195              & \cellcolor{blue!20}0.137             & \cellcolor{blue!10}1.179          & \cellcolor{blue!20}0.540           \\ \hline
\end{tabular}
}
\caption{Results of temperature scaling for one of the \OSM{} -  \texttt{LLaMA2-70B}. The darker \colorbox{blue!20}{blue} shade highlights better performing prompting technique. }
\label{table : temp}
\end{table*}

\paragraph{Better calibration observed for \pal{} when inducing similarity in generations for \texttt{LLaMA2-70B}} 
We observe that for \CSM{}, \pal{} is not only more accurate but also more calibrated than \COT{}. 
Consequently, we explore whether the reduction in generation diversity, achievable through lower temperatures, can contribute to improved calibration for \OSM{}.


We perform a parameter sweep across temperature values ranging between 0.1 and 0.7 with a step size of 0.2.  
We show the variation of accuracy, calibration, generation similarity, and answer entropy for two datasets in Figure \ref{fig:temp}. The plots for the remaining datasets are available in Appendix \ref{sec:B}, Figure \ref{fig:temp-app}. We can see that we obtain better calibration in \texttt{LLaMA2-70B} in both \pal{} and \COT{} for temperatures below 1.0.  From Table \ref{table : temp} we note that in the majority of runs with T < 1.0, \pal{} is better calibrated than \COT{}. Optimal performance, considering accuracy and calibration, is achieved at different temperatures for each dataset. For most T values, we note that the similarity scores are higher while corresponding answer entropy values are lower for \pal{} compared to \COT{}. This mirrors the pattern observed for \CSM{}.

However, optimal temperature values in our runs for calibration are either 0.5 or 0.7, while extreme values (0.1, 1.0)  yield lower calibration and accuracy performance. We can therefore see that scaling temperatures in the \OSM{} can help us to obtain better calibration for \pal{}, which already performs better than \COT{} on these reasoning tasks. 

Overall, we see that lower generation diversity and lower answer entropy lead to higher calibration up to a certain point, after which it negatively affects the calibration. (\textit{\textbf{--RQ3}}) 

\begin{figure*}[h]
    \centering

    \begin{minipage}{0.5\textwidth}
        \centering
        \includegraphics[width=\linewidth]{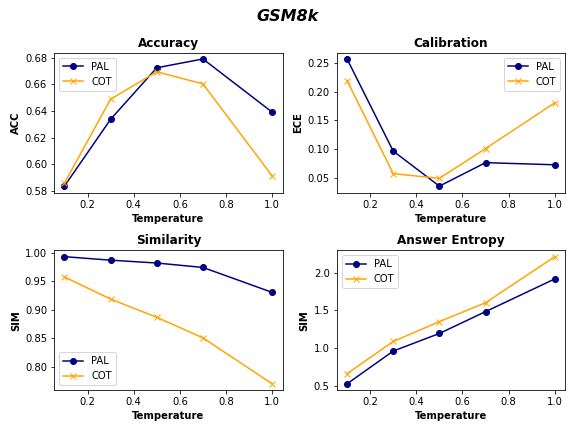}
    \end{minipage}%
    \begin{minipage}{0.5\textwidth}
        \centering
        \includegraphics[width=\linewidth]{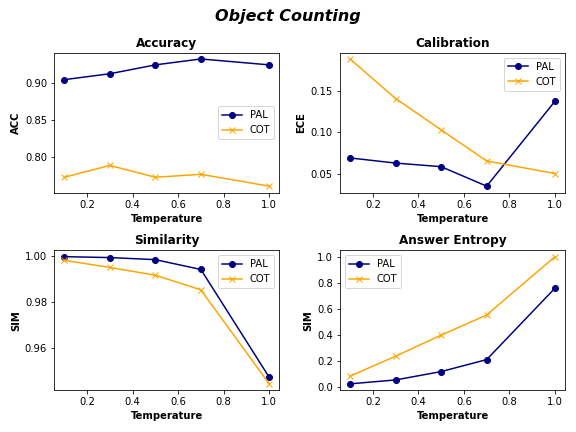}
    \end{minipage}

    \caption{Trends seen in temperature scaling for the model \texttt{LLaMA2-70B}. Across datasets, the accuracy and calibration improve upon lower the temperature up to a certain extent. This is in line with having lower generation similarity and lower answer entropy. The optimal temperatures seen are 0.5 and 0.7 across datasets. For other datasets, refer Appendix, Figure \ref{fig:temp-app}.}
    \label{fig:temp}
\end{figure*}

\section{Related Work}

\subsection{Prompting Strategies for Reasoning}
Recent developments in language models have introduced various methods to enhance their reasoning abilities. One such method is  CoT \cite{Wei2022ChainOT} which helps models generate a series of intermediate steps to solve problems. CoT has demonstrated improved performance in tasks involving arithmetic, common sense, and symbolic reasoning. There are approaches such as PaL \cite{Gao2022PALPL} and Program-of-thoughts (PoT)  \cite{Chen2022ProgramOT}  which go a step further by generating programs as intermediate steps and using an interpreter to process them.
Code as a medium of reasoning has shown considerable promise evidenced by better performance over chain-of-thought style prompting strategies, in several recent studies~\cite{Madaan2022LanguageMO, Gao2022PALPL, lyu2023faithful, Zhang2023ExploringTC, Zhang2023CausalRO}.
Different from these works, our main goal in this paper is to understand the effect of code prompts on calibration.

\vspace{-0.3em}
\subsection{Calibration in Language Models}
Calibration has been extensively studied in structured prediction problems, such as named entity recognition and part of speech tagging \cite{jagannatha2020calibrating}, as well as in natural language understanding tasks, like question answering and text classification \cite{kamath2020selective, kong2020calibrated, desai2020calibration}. More recently, studies have directed their attention to calibrating language models when used as generators \cite{jiang2021can, zhao2021calibrate}.  Additionally, the study by \citet{kadavath2022language} explored the likelihood of a model knowing the answer before proposing a response. However, all of these approaches typically rely on access to the model's logits.

In contrast, the work by \cite{tian2023just} investigates verbalized probability estimates to assess the calibration of large language models without needing access to logits. This involves the practice of querying the model about its confidence in the answers it generates. Furthermore, \cite{xiong2023can} introduced self-consistency-based methods for calibration, demonstrating their superior performance compared to verbalized methods. In our research, we adopt self-consistency as the method of choice for measuring calibration.

\section{Conclusion}

In this study, we explore the impact of two distinct prompting styles, namely \pal{} and \COT{}, on the calibration of \CSM{} and \OSM{}. Our investigation spans 5 reasoning datasets, employing self-consistency as the methodology for eliciting calibration. 
We analyze four different metrics - calibration (ECE) , accuracy (ACC) , average similarity in generations (SIM) , and answer entropy (ENT) . We summarize our findings as follows: 

\begin{itemize}

    \item \textit{ \textbf{RQ 1:} Does one prompting style result in significantly better calibration than the other?}
    Empirical results show that \pal{} generally has higher calibration and accuracy for 82.5\% of the cases across OpenAI and LLaMA models for a varied range of temperatures.

    \item \textit{ \textbf{RQ 2:} Are the observed calibration trends different across  \CSM{} and \OSM{}?}
    We observed that \CSM{} are in general better calibrated for the reasoning tasks with up to 19\% improvement in ECE score.
    
    \item \textit{\textbf{RQ 3:} Does the consistency of LLM generations affect performance? } 
    \pal{} prompting shows a general trend of having greater similarity in the generation of up to 7\% over text, which we hypothesize could be due to the inherent structure present in the code. We see that greater generation similarity is accompanied by lower answer entropy and lower ECE. However, temperature scaling experiments reveal that reducing generation diversity helps improve calibration only up to certain temperature values -- the calibration is affected negatively for lower temperatures such as 0.1 and 0.3.
\end{itemize}


We hope that this study will serve as a catalyst for additional research aimed at holistically evaluating and gaining deeper insights into the role of prompts in various task domains across other dimensions in addition to accuracy.

\section{Acknowledgments}

This work was supported by an NEC Student Research Fellowship and a PGS-D fellowship from the Natural Sciences and Engineering Research Council of Canada (NSERC), [award number 578085-2023]

\bibliography{custom}

\begin{thebibliography}{33}
\expandafter\ifx\csname natexlab\endcsname\relax\def\natexlab#1{#1}\fi

\bibitem[{Brown et~al.(2020)Brown, Mann, Ryder, Subbiah, Kaplan, Dhariwal,
  Neelakantan, Shyam, Sastry, Askell et~al.}]{brown2020language}
Tom Brown, Benjamin Mann, Nick Ryder, Melanie Subbiah, Jared~D Kaplan, Prafulla
  Dhariwal, Arvind Neelakantan, Pranav Shyam, Girish Sastry, Amanda Askell,
  et~al. 2020.
\newblock Language models are few-shot learners.
\newblock \emph{Advances in neural information processing systems},
  33:1877--1901.

\bibitem[{Chen et~al.(2022)Chen, Ma, Wang, and Cohen}]{Chen2022ProgramOT}
Wenhu Chen, Xueguang Ma, Xinyi Wang, and William~W. Cohen. 2022.
\newblock Program of thoughts prompting: Disentangling computation from
  reasoning for numerical reasoning tasks.
\newblock \emph{ArXiv}, abs/2211.12588.

\bibitem[{Cobbe et~al.(2021)Cobbe, Kosaraju, Bavarian, Hilton, Nakano, Hesse,
  and Schulman}]{Cobbe2021TrainingVT}
Karl Cobbe, Vineet Kosaraju, Mohammad Bavarian, Jacob Hilton, Reiichiro Nakano,
  Christopher Hesse, and John Schulman. 2021.
\newblock Training verifiers to solve math word problems.
\newblock \emph{ArXiv}, abs/2110.14168.

\bibitem[{Desai and Durrett(2020)}]{desai2020calibration}
Shrey Desai and Greg Durrett. 2020.
\newblock Calibration of pre-trained transformers.
\newblock \emph{arXiv preprint arXiv:2003.07892}.

\bibitem[{Gao et~al.(2022)Gao, Madaan, Zhou, Alon, Liu, Yang, Callan, and
  Neubig}]{Gao2022PALPL}
Luyu Gao, Aman Madaan, Shuyan Zhou, Uri Alon, Pengfei Liu, Yiming Yang, Jamie
  Callan, and Graham Neubig. 2022.
\newblock Pal: Program-aided language models.
\newblock \emph{ArXiv}, abs/2211.10435.

\bibitem[{Guo et~al.(2017{\natexlab{a}})Guo, Pleiss, Sun, and
  Weinberger}]{guo2017calibration}
Chuan Guo, Geoff Pleiss, Yu~Sun, and Kilian~Q Weinberger. 2017{\natexlab{a}}.
\newblock On calibration of modern neural networks.
\newblock In \emph{International conference on machine learning}, pages
  1321--1330. PMLR.

\bibitem[{Guo et~al.(2017{\natexlab{b}})Guo, Pleiss, Sun, and
  Weinberger}]{Guo2017OnCO}
Chuan Guo, Geoff Pleiss, Yu~Sun, and Kilian~Q. Weinberger. 2017{\natexlab{b}}.
\newblock On calibration of modern neural networks.
\newblock In \emph{International Conference on Machine Learning}.

\bibitem[{Holtzman et~al.(2020)Holtzman, Buys, Du, Forbes, and
  Choi}]{holtzman2020curious}
Ari Holtzman, Jan Buys, Li~Du, Maxwell Forbes, and Yejin Choi. 2020.
\newblock \href {http://arxiv.org/abs/1904.09751} {The curious case of neural
  text degeneration}.

\bibitem[{Jagannatha and Yu(2020)}]{jagannatha2020calibrating}
Abhyuday Jagannatha and Hong Yu. 2020.
\newblock Calibrating structured output predictors for natural language
  processing.
\newblock In \emph{Proceedings of the conference. Association for Computational
  Linguistics. Meeting}, volume 2020, page 2078. NIH Public Access.

\bibitem[{Jiang et~al.(2020)Jiang, Araki, Ding, and Neubig}]{Jiang2020HowCW}
Zhengbao Jiang, J.~Araki, Haibo Ding, and Graham Neubig. 2020.
\newblock How can we know when language models know? on the calibration of
  language models for question answering.
\newblock \emph{Transactions of the Association for Computational Linguistics},
  9:962--977.

\bibitem[{Jiang et~al.(2021)Jiang, Araki, Ding, and Neubig}]{jiang2021can}
Zhengbao Jiang, Jun Araki, Haibo Ding, and Graham Neubig. 2021.
\newblock How can we know when language models know? on the calibration of
  language models for question answering.
\newblock \emph{Transactions of the Association for Computational Linguistics},
  9:962--977.

\bibitem[{Kadavath et~al.(2022)Kadavath, Conerly, Askell, Henighan, Drain,
  Perez, Schiefer, Hatfield-Dodds, DasSarma, Tran-Johnson
  et~al.}]{kadavath2022language}
Saurav Kadavath, Tom Conerly, Amanda Askell, Tom Henighan, Dawn Drain, Ethan
  Perez, Nicholas Schiefer, Zac Hatfield-Dodds, Nova DasSarma, Eli
  Tran-Johnson, et~al. 2022.
\newblock Language models (mostly) know what they know.
\newblock \emph{arXiv preprint arXiv:2207.05221}.

\bibitem[{Kamath et~al.(2020)Kamath, Jia, and Liang}]{kamath2020selective}
Amita Kamath, Robin Jia, and Percy Liang. 2020.
\newblock Selective question answering under domain shift.
\newblock \emph{arXiv preprint arXiv:2006.09462}.

\bibitem[{Kong et~al.(2020)Kong, Jiang, Zhuang, Lyu, Zhao, and
  Zhang}]{kong2020calibrated}
Lingkai Kong, Haoming Jiang, Yuchen Zhuang, Jie Lyu, Tuo Zhao, and Chao Zhang.
  2020.
\newblock Calibrated language model fine-tuning for in-and out-of-distribution
  data.
\newblock \emph{arXiv preprint arXiv:2010.11506}.

\bibitem[{Li et~al.(2022)Li, Choi, Chung, Kushman, Schrittwieser, Leblond,
  Eccles, Keeling, Gimeno, Dal~Lago et~al.}]{li2022competition}
Yujia Li, David Choi, Junyoung Chung, Nate Kushman, Julian Schrittwieser,
  R{\'e}mi Leblond, Tom Eccles, James Keeling, Felix Gimeno, Agustin Dal~Lago,
  et~al. 2022.
\newblock Competition-level code generation with alphacode.
\newblock \emph{Science}, 378(6624):1092--1097.

\bibitem[{Lyu et~al.(2023)Lyu, Havaldar, Stein, Zhang, Rao, Wong, Apidianaki,
  and Callison-Burch}]{lyu2023faithful}
Qing Lyu, Shreya Havaldar, Adam Stein, Li~Zhang, Delip Rao, Eric Wong, Marianna
  Apidianaki, and Chris Callison-Burch. 2023.
\newblock Faithful chain-of-thought reasoning.
\newblock \emph{arXiv preprint arXiv:2301.13379}.

\bibitem[{Madaan et~al.(2022)Madaan, Zhou, Alon, Yang, and
  Neubig}]{Madaan2022LanguageMO}
Aman Madaan, Shuyan Zhou, Uri Alon, Yiming Yang, and Graham Neubig. 2022.
\newblock Language models of code are few-shot commonsense learners.
\newblock \emph{ArXiv}, abs/2210.07128.

\bibitem[{McLean et~al.(1991)McLean, Sanders, and Stroup}]{mclean1991unified}
Robert~A McLean, William~L Sanders, and Walter~W Stroup. 1991.
\newblock A unified approach to mixed linear models.
\newblock \emph{The American Statistician}, 45(1):54--64.

\bibitem[{{OpenAI}(2023)}]{openAI}
{OpenAI}. 2023.
\newblock Openai documentation.
\newblock \url{https://platform.openai.com/docs/model-index-for-researchers}.

\bibitem[{Platt et~al.(1999)}]{platt1999probabilistic}
John Platt et~al. 1999.
\newblock Probabilistic outputs for support vector machines and comparisons to
  regularized likelihood methods.
\newblock \emph{Advances in large margin classifiers}, 10(3):61--74.

\bibitem[{Suzgun et~al.(2022{\natexlab{a}})Suzgun, Scales, Sch{\"a}rli,
  Gehrmann, Tay, Chung, Chowdhery, Le, Chi, Zhou
  et~al.}]{suzgun2022challenging}
Mirac Suzgun, Nathan Scales, Nathanael Sch{\"a}rli, Sebastian Gehrmann, Yi~Tay,
  Hyung~Won Chung, Aakanksha Chowdhery, Quoc~V Le, Ed~H Chi, Denny Zhou, et~al.
  2022{\natexlab{a}}.
\newblock Challenging big-bench tasks and whether chain-of-thought can solve
  them.
\newblock \emph{arXiv preprint arXiv:2210.09261}.

\bibitem[{Suzgun et~al.(2022{\natexlab{b}})Suzgun, Scales, Scharli, Gehrmann,
  Tay, Chung, Chowdhery, Le, hsin Chi, Zhou, and Wei}]{Suzgun2022ChallengingBT}
Mirac Suzgun, Nathan Scales, Nathanael Scharli, Sebastian Gehrmann, Yi~Tay,
  Hyung~Won Chung, Aakanksha Chowdhery, Quoc~V. Le, Ed~Huai hsin Chi, Denny
  Zhou, and Jason Wei. 2022{\natexlab{b}}.
\newblock Challenging big-bench tasks and whether chain-of-thought can solve
  them.
\newblock \emph{ArXiv}, abs/2210.09261.

\bibitem[{Tian et~al.(2023)Tian, Mitchell, Zhou, Sharma, Rafailov, Yao, Finn,
  and Manning}]{tian2023just}
Katherine Tian, Eric Mitchell, Allan Zhou, Archit Sharma, Rafael Rafailov,
  Huaxiu Yao, Chelsea Finn, and Christopher~D Manning. 2023.
\newblock Just ask for calibration: Strategies for eliciting calibrated
  confidence scores from language models fine-tuned with human feedback.
\newblock \emph{arXiv preprint arXiv:2305.14975}.

\bibitem[{Touvron et~al.(2023)Touvron, Martin, Stone, Albert, Almahairi,
  Babaei, Bashlykov, Batra, Bhargava, Bhosale et~al.}]{touvron2023llama}
Hugo Touvron, Louis Martin, Kevin Stone, Peter Albert, Amjad Almahairi, Yasmine
  Babaei, Nikolay Bashlykov, Soumya Batra, Prajjwal Bhargava, Shruti Bhosale,
  et~al. 2023.
\newblock Llama 2: Open foundation and fine-tuned chat models.
\newblock \emph{arXiv preprint arXiv:2307.09288}.

\bibitem[{Wang et~al.(2022)Wang, Wei, Schuurmans, Le, hsin Chi, and
  Zhou}]{Wang2022SelfConsistencyIC}
Xuezhi Wang, Jason Wei, Dale Schuurmans, Quoc Le, Ed~Huai hsin Chi, and Denny
  Zhou. 2022.
\newblock Self-consistency improves chain of thought reasoning in language
  models.
\newblock \emph{ArXiv}, abs/2203.11171.

\bibitem[{Wei et~al.(2022)Wei, Wang, Schuurmans, Bosma, hsin Chi, Xia, Le, and
  Zhou}]{Wei2022ChainOT}
Jason Wei, Xuezhi Wang, Dale Schuurmans, Maarten Bosma, Ed~Huai hsin Chi,
  F.~Xia, Quoc Le, and Denny Zhou. 2022.
\newblock Chain of thought prompting elicits reasoning in large language
  models.
\newblock \emph{ArXiv}, abs/2201.11903.

\bibitem[{Xiong et~al.(2023{\natexlab{a}})Xiong, Hu, Lu, Li, Fu, He, and
  Hooi}]{Xiong2023CanLE}
Miao Xiong, Zhiyuan Hu, Xinyang Lu, Yifei Li, Jie Fu, Junxian He, and Bryan
  Hooi. 2023{\natexlab{a}}.
\newblock Can llms express their uncertainty? an empirical evaluation of
  confidence elicitation in llms.
\newblock \emph{ArXiv}, abs/2306.13063.

\bibitem[{Xiong et~al.(2023{\natexlab{b}})Xiong, Hu, Lu, Li, Fu, He, and
  Hooi}]{xiong2023can}
Miao Xiong, Zhiyuan Hu, Xinyang Lu, Yifei Li, Jie Fu, Junxian He, and Bryan
  Hooi. 2023{\natexlab{b}}.
\newblock Can llms express their uncertainty? an empirical evaluation of
  confidence elicitation in llms.
\newblock \emph{arXiv preprint arXiv:2306.13063}.

\bibitem[{Yao et~al.(2023)Yao, Yu, Zhao, Shafran, Griffiths, Cao, and
  Narasimhan}]{Yao2023TreeOT}
Shunyu Yao, Dian Yu, Jeffrey Zhao, Izhak Shafran, Thomas~L. Griffiths, Yuan
  Cao, and Karthik Narasimhan. 2023.
\newblock Tree of thoughts: Deliberate problem solving with large language
  models.
\newblock \emph{ArXiv}, abs/2305.10601.

\bibitem[{Zhang et~al.(2023{\natexlab{a}})Zhang, Dugan, Xu, and
  Callison-Burch}]{Zhang2023ExploringTC}
Li~Zhang, Liam Dugan, Hai Xu, and Chris Callison-Burch. 2023{\natexlab{a}}.
\newblock Exploring the curious case of code prompts.
\newblock \emph{ArXiv}, abs/2304.13250.

\bibitem[{Zhang et~al.(2023{\natexlab{b}})Zhang, Xu, Yang, Zhou, You, Arora,
  and Callison-Burch}]{Zhang2023CausalRO}
Li~Zhang, Hai Xu, Yue Yang, Shuyan Zhou, Weiqiu You, Manni Arora, and Chris
  Callison-Burch. 2023{\natexlab{b}}.
\newblock Causal reasoning of entities and events in procedural texts.
\newblock In \emph{Findings}.

\bibitem[{Zhao et~al.(2021)Zhao, Wallace, Feng, Klein, and
  Singh}]{zhao2021calibrate}
Zihao Zhao, Eric Wallace, Shi Feng, Dan Klein, and Sameer Singh. 2021.
\newblock Calibrate before use: Improving few-shot performance of language
  models.
\newblock In \emph{International Conference on Machine Learning}, pages
  12697--12706. PMLR.

\bibitem[{Zhou et~al.(2022)Zhou, Scharli, Hou, Wei, Scales, Wang, Schuurmans,
  Bousquet, Le, and hsin Chi}]{Zhou2022LeasttoMostPE}
Denny Zhou, Nathanael Scharli, Le~Hou, Jason Wei, Nathan Scales, Xuezhi Wang,
  Dale Schuurmans, Olivier Bousquet, Quoc Le, and Ed~Huai hsin Chi. 2022.
\newblock Least-to-most prompting enables complex reasoning in large language
  models.
\newblock \emph{ArXiv}, abs/2205.10625.

\end{thebibliography}
\bibliographystyle{acl_natbib}
\clearpage
\appendix

\usemintedstyle{xcode}
\onecolumn




  

\section{Prompts}
\label{sec:prompts}
The following sections display one example of the few-shot prompts used for each dataset across prompting styles.
\subsection{\pal{} Prompts}
\subsubsection{GSM8K/GSM8K-Hard}
\begin{tcolorbox}[width=\textwidth, boxrule=1pt, colback=white, arc=1pt, outer arc=1pt]
\begin{minted}[linenos=false, fontsize=\small]{python}
def solution ()  :
    """Olivia has $23. She bought five bagels for $3 each. How much money does she have left?"""
    money_initial = 23
    bagels = 5
    bagel_cost = 3
    money_spent = bagels * bagel_cost
    money_left = money_initial - money_spent
    result = money_left
    return result
\end{minted}
\end{tcolorbox}

\subsubsection{Object Counting}

\begin{tcolorbox}[width=\textwidth, boxrule=1pt, colback=white, arc=1pt, outer arc=1pt]
\begin{minted}[linenos=false, fontsize=\small, breaklines]{python}
# Q: I have a chair, two potatoes, a cauliflower, a lettuce head, two tables, a cabbage, two onions, and three fridges. How many vegetables do I have?
```
def solution ()  :
	# note: I'm not counting the chair, tables, or fridges
	vegetables_to_count = {{'potato': 2,'cauliflower': 1,'lettuce head': 1,'cabbage': 1,'onion': 2}}
	return sum (vegetables_to_count.values () )  
```
\end{minted}
\end{tcolorbox}

\subsubsection{Date Understanding}

\begin{tcolorbox}[width=\textwidth, boxrule=1pt, colback=white, arc=1pt, outer arc=1pt]
\begin{minted}[linenos=false, fontsize=\small, breaklines]{python}
# Q: 2015 is coming in 36 hours. What is the date one week from today in MM/DD/YYYY?
# If 2015 is coming in 36 hours, then today is 36 hours before.
today = datetime (2015, 1, 1)  - relativedelta (hours=36) 
# One week from today,
one_week_from_today = today + relativedelta (weeks=1) 
# The answer formatted with %m/%d/%Y is
one_week_from_today.strftime ('%m/%d/%Y') 
\end{minted}
\end{tcolorbox}

\subsubsection{Repeat Copy}

\begin{tcolorbox}[width=\textwidth, boxrule=1pt, colback=white, arc=1pt, outer arc=1pt]
\begin{minted}[linenos=false, fontsize=\small, breaklines]{python}
# Q: Repeat the word duck four times, but halfway through also say quack
```
def solution ()  :
	result = []
	for i in range (1, 5) :
		result.append ("duck") 
		if i == 2:
			result.append ("quack") 
	return " ".join (result) 
```
\end{minted}
\end{tcolorbox}

\clearpage
\newpage
\subsection{\COT{} Prompts}

\subsubsection{GSM8K/GSM8K-Hard}
\begin{tcolorbox}[width=\textwidth, boxrule=1pt, colback=white, arc=1pt, outer arc=1pt, fontupper=\small, fontlower=\small]
\textbf{Q}: There are 15 trees in the grove. Grove workers will plant trees in the grove today. After they are done, there will be 21 trees. How many trees did the grove workers plant today?

\textbf{A}: There are 15 trees originally. Then there were 21 trees after some more were planted. So there must have been 21 - 15 = 6. The answer is 6.
\end{tcolorbox}

\subsubsection{Object Counting}
\begin{tcolorbox}[width=\textwidth, boxrule=1pt, colback=white, arc=1pt, outer arc=1pt, fontupper=\small, fontlower=\small]
\textbf{Q}: I have a chair, two potatoes, a cauliflower, a lettuce head, two tables, a cabbage, two onions, and three fridges. How many vegetables do I have?

\textbf{A}: 
Chair, tables and fridges are not vegetables, so we are not counting them.
Two potatoes, cauliflower, lettuce head, cabbage and two onions are vegetables, so they will be counted
The total number of vegetables are 2 + 1 + 1 + 1 + 2
The answer is: 7
\end{tcolorbox}

\subsubsection{Date Understanding}

\begin{tcolorbox}[width=\textwidth, boxrule=1pt, colback=white, arc=1pt, outer arc=1pt, fontupper=\small, fontlower=\small]
\textbf{Q}: 2015 is coming in 36 hours. What is the date one week from today in MM/DD/YYYY?

\textbf{A}: If 2015 is coming in 36 hours, then it is coming in 2 days. 2 days before 01/01/2015 is 12/30/2014, so today is 12/30/2014. So one week from today will be 01/05/2015. So the answer is 01/05/2015.
\end{tcolorbox}

\subsubsection{Repeat Copy}

\begin{tcolorbox}[width=\textwidth, boxrule=1pt, colback=white, arc=1pt, outer arc=1pt, fontupper=\small, fontlower=\small]
\textbf{Q}: Repeat the word duck four times, but halfway through also say quack

\textbf{A}:
On repeating the word duck four times we get: duck duck duck duck
Halfway through if we say quack, we have to say quack in between the 2nd word and the 3rd word
The answer is: duck duck quack duck duck
\end{tcolorbox}
\clearpage

\section{Temperature Scaling Experiments - Line Plots}
\label{sec:B}
\FloatBarrier
\begin{figure*}[h]
    \centering

    \begin{minipage}{0.5\textwidth}
        \centering
        \includegraphics[width=\linewidth]{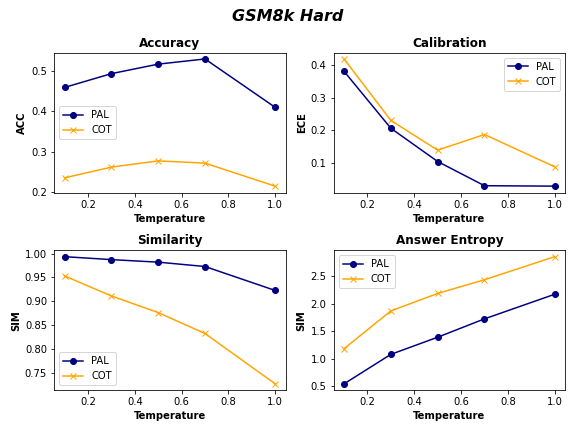}
    \end{minipage}%
    \begin{minipage}{0.5\textwidth}
        \centering
        \includegraphics[width=\linewidth]{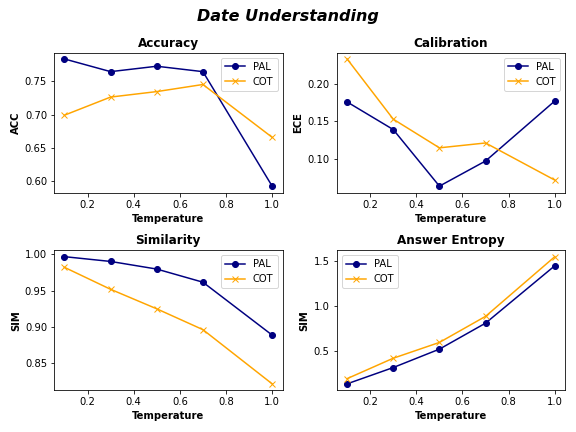}
    \end{minipage}
    \begin{minipage}{0.5\textwidth}
        \centering
        \includegraphics[width=\linewidth]{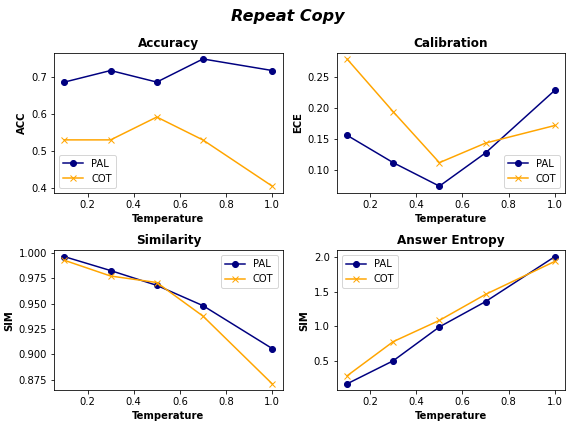}
    \end{minipage}
    \caption{Trends seen in temperature scaling for GSM8K Hard, Date Understanding and Repeat Copy}
    \label{fig:temp-app}
\end{figure*}

\end{document}